\documentclass[10pt,twocolumn,letterpaper]{article}

\usepackage[pagenumbers]{cvpr}

\usepackage[dvipsnames]{xcolor}
\usepackage{graphicx}
\usepackage{amsmath}
\usepackage{amssymb}
\usepackage{booktabs}
\usepackage{bm}
\usepackage{bbm}
\usepackage{siunitx}
\usepackage{multirow}

\newcommand{\dashrule}[1][black]{
  \color{#1}\rule[\dimexpr.5ex-.2pt]{4pt}{.4pt}\xleaders\hbox{\rule{4pt}{0pt}\rule[\dimexpr.5ex-.2pt]{4pt}{.4pt}}\hfill\kern0pt
}

\DeclareMathOperator*{\argmin}{arg\,min}

\definecolor{cvprblue}{rgb}{0.21,0.49,0.74}
\usepackage[pagebackref,breaklinks,colorlinks,citecolor=cvprblue]{hyperref}

\def\paperID{}
\def\confName{}
\def\confYear{}

\title{Towards Precise Action Spotting: Addressing Temporal Misalignment in Labels with Dynamic Label Assignment}

\author{Masato Tamura\\
Hitachi America, Ltd.\\
2535 Augustine Dr, Santa Clara, California, United States of America, 95054\\
{\tt\small masato.tamura@ieee.org}
}

\begin{document}
\maketitle
\begin{abstract}
Precise action spotting has attracted considerable attention due to its promising applications. While existing methods achieve substantial performance by employing well-designed model architecture, they overlook a significant challenge: the temporal misalignment inherent in ground-truth labels. This misalignment arises when frames labeled as containing events do not align accurately with the actual event times, often as a result of human annotation errors or the inherent difficulties in precisely identifying event boundaries across neighboring frames. To tackle this issue, we propose a novel dynamic label assignment strategy that allows predictions to have temporal offsets from ground-truth action times during training, ensuring consistent event spotting. Our method extends the concept of minimum-cost matching, which is utilized in the spatial domain for object detection, to the temporal domain. By calculating matching costs based on predicted action class scores and temporal offsets, our method dynamically assigns labels to the most likely predictions, even when the predicted times of these predictions deviate from ground-truth times, alleviating the negative effects of temporal misalignment in labels. We conduct extensive experiments and demonstrate that our method achieves state-of-the-art performance, particularly in conditions where events are visually distinct and temporal misalignment in labels is common.
\end{abstract}

\vspace{-1em}    
\section{Introduction}
\label{sec:intro}

Action understanding in videos has been extensively researched for its potential applications, such as sports analysis, surveillance, and human-computer interaction. The scope of action understanding tasks extends from basic action recognition in trimmed videos~\cite{karpathy_cvpr2014,du_iccv2015,joao_cvpr2017}, which involves classifying the types of actions depicted in pre-segmented clips, to more sophisticated tasks such as spatiotemporal action localization in untrimmed videos~\cite{weinzaepfel_iccv2015,gritsenko_cvpr2024,li_eccv2020,pan_cvpr2021}, where models localize and classify actions by identifying actors and their actions over time with bounding boxes. Despite recent advances, the need for robust and efficient solutions for real-world applications keeps action understanding an active and evolving research topic.

\begin{figure}[t]
    \centering
    \begin{subfigure}{\linewidth}
        \includegraphics[keepaspectratio,width=\linewidth]{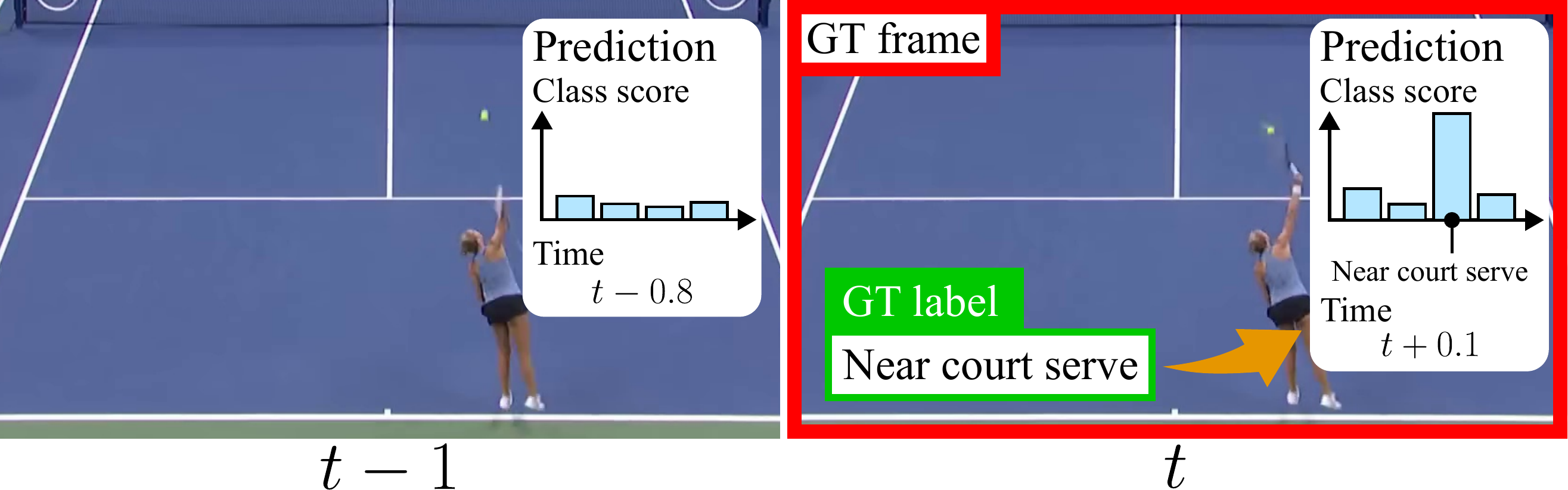}
        \caption{Ground-truth label without temporal misalignment.}
        \label{fig:dynamic_precise}
    \end{subfigure} \\
    \begin{subfigure}{\linewidth}
        \includegraphics[keepaspectratio,width=\linewidth]{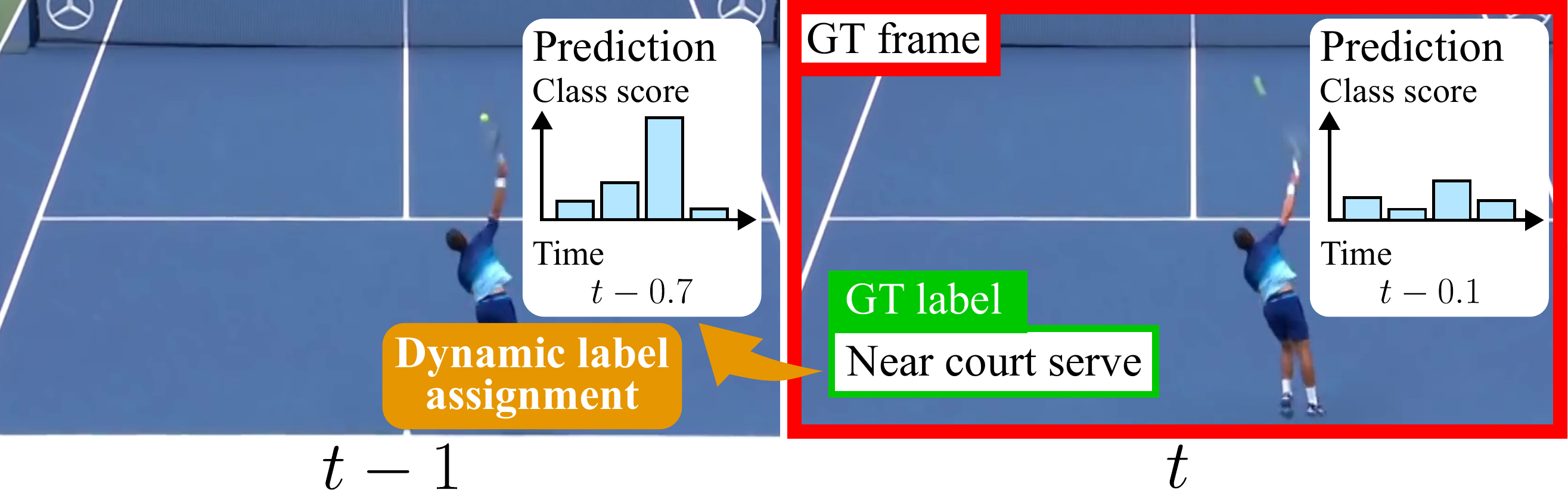}
        \caption{Ground-truth label with temporal misalignment.}
        \label{fig:dynamic_gap}
    \end{subfigure}
    \caption{Two examples of our proposed dynamic label assignment. In both cases, the action label ``Near court serve" is tagged to the frames at time $t$. In example (a), the event appears to occur around $t$, leading our method to assign the label to the prediction with a predicted time close to $t$ during training. Conversely, in example (b), the event appears to take place around $t - 1$. In this case, our method assigns the label to the prediction with a predicted time close to $t - 1$ based on the predicted action class scores.}
    \label{fig:dynamic}
    \vspace{-1.5em}
\end{figure}


To further expand the capability of action understanding in real-world scenarios, precise action spotting has been proposed~\cite{hong_eccv2022}. In the precise action spotting task, models aim to precisely identify frames where specific events occur. For example, in video clips of tennis matches, models are required to identify frames where players serve at the near court locations, as shown in Fig~\ref{fig:dynamic}. Since adjacent frames typically look similar, it is challenging to identify event frames precisely. To tackle this issue, existing methods~\cite{hong_eccv2022,xarles_cvprw2024} have designed model architectures that effectively consider subtle frame differences during the early stages of feature extraction and demonstrate substantial performance in benchmark datasets.


While we leverage existing model designs, we approach the challenge of precise action spotting from a distinct perspective: addressing the issue of temporal misalignment in ground-truth labels. Figure~\ref{fig:dynamic_gap} illustrates an example of the misalignment. In the example, the frame at time $t$ is tagged with the action label, yet the event appears to occur around $t - 1$ based on the appearance of the frame at $t - 1$. This temporal misalignment can result not only from human annotation errors but also from the inherent difficulties of accurately identifying event boundaries across neighboring frames. Furthermore, precisely annotating event frames can be cumbersome, as annotators need to repeatedly navigate back and forth between frames. By allowing for a certain degree of misalignment between event and tagged frames, the annotation process can be more efficient, thereby making precise action spotting more practical.


To address the issue, we propose a dynamic label assignment strategy that effectively assigns ground-truth labels to the most plausible predictions during training. This method draws inspiration from the recent advancements in object detection, particularly the end-to-end methodology introduced by Carion~\etal~\cite{carion_eccv2020}.
Their framework calculates matching costs between predicted and ground-truth bounding boxes and subsequently identifies optimal matching pairs that yield the lowest total cost.
Our method extends the minimum cost matching from the spatial domain into the temporal domain. Specifically, our method computes matching costs using predicted action class scores and temporal offsets from the times of tagged frames and identifies matching pairs that minimize the total cost. Notably, while our dynamic label assignment method serves to alleviate the need for cumbersome post-processing, as seen in object detection models~\cite{carion_eccv2020}, our primary objective is to solve the issue of temporal misalignment in labels. By employing predicted action class scores in the matching process, our method enables labels to be assigned to predictions whose predicted times have some offsets from the times of tagged frames. Figure~\ref{fig:dynamic} shows two examples of our dynamic label assignment. In both examples, the action label ``Near court serve" is tagged to the frames at time $t$. When the event timing is accurately aligned with the label, our method assigns the label to the prediction whose predicted time is close to $t$, as shown in Fig.~\ref{fig:dynamic_precise}. Conversely, when the event appears to occur around $t - 1$, our method assigns the label to the prediction close to $t - 1$ based on the predicted action class scores that demonstrate stronger alignment with the action label than the scores of the prediction close to $t$, as depicted in Fig.~\ref{fig:dynamic_gap}. This dynamic label assignment mitigates the challenges of inconsistent action spotting posed by temporal misalignment in labels, thereby enabling models to learn precise action spotting effectively.


In summary, our contributions are three-fold: (1) We approach the issue of precise action spotting from a novel perspective that distinguishes our work from existing research. To our knowledge, this is the first study that addresses temporal misalignment in ground-truth labels for precise action spotting. (2) Our method achieves state-of-the-art performance on multiple benchmark datasets. In particular, our method is effective in scenarios where events are visually distinct and temporal misalignment in labels is common. (3) We conduct extensive experiments to verify that our method successfully mitigates the effects of temporal misalignment in labels through our dynamic label assignment.
\section{Related work}
\label{sec:related_work}

\subsection{Action understanding in videos}


The tasks associated with action understanding have grown increasingly intricate, where models are required to identify the temporal beginnings and endings of actions within untrimmed videos~\cite{xu_iccv2017,chao_cvpr2018,lin_cvpr2021,zhang_eccv2022}, identify actors involved in those actions in addition to the temporal timings~\cite{weinzaepfel_iccv2015,gritsenko_cvpr2024,li_eccv2020,pan_cvpr2021}, or temporally segment untrimmed videos based on pre-defined action categories~\cite{lu_cvpr2024,yi_bmvc2021,lei_cvpr2018,gong_nips2023}. Since all these tasks require models to determine the temporal intervals of actions, feature extraction across frames is essential. The attention mechanism in transformers~\cite{vaswani_nips2017} plays a pivotal role for this feature extraction. Zhang~\etal~\cite{zhang_eccv2022} proposed multi-scale transformers to identify actions of varying temporal lengths.
Yi~\etal~\cite{yi_bmvc2021} introduced heuristic biases into transformers to enhance training efficiency. In our work, we also leverage transformers for our dynamic label assignment.


To broaden the scope of action understanding, Hong~\etal~\cite{hong_eccv2022} introduced the concept of precise action spotting. They developed a spatiotemporal feature extraction model designed to capture subtle visual differences and created task datasets to advance research in this area. Building on this foundation, Xarles~\etal~\cite{xarles_cvprw2024} enhanced model designs to further improve performance. Instead of focusing solely on enhancing feature extraction capabilities, we propose a precise action-spotting method aimed at reducing the impact of temporal misalignment in labels.

\subsection{Dynamic label assignment in object detection}

Dynamic label assignment was initially introduced in a transformer-based object detector known as DETR~\cite{carion_eccv2020}. DETR dynamically assigns each ground-truth box to the most plausible predicted box based on predicted class scores and geometric values of bounding boxes, enabling the model to learn to output a single box for each object and thus supporting end-to-end object detection. Since then, this strategy has been incorporated into convolutional neural network (CNN)-based object detectors~\cite{feng_iccv2021,ge_arxiv2021,ge_cvpr2021,lyu_arxiv2022} as well. While these approaches do not focus on end-to-end object detection, they aim to improve convergence speed and model performance through one-to-many assignments.

In our work, we leverage the dynamic label assignment strategy in DETR not to train end-to-end models but to address the challenges of inconsistent action spotting posed by temporal misalignment in labels. By calculating matching costs with predicted action class scores, this strategy assigns ground-truth labels to the most plausible predictions, thereby facilitating consistent precise action spotting.

\section{Proposed method}
\label{sec:proposed_method}

Inspired by DETR~\cite{carion_eccv2020}, we leverage transformers~\cite{vaswani_nips2017} to dynamically assign ground truth labels to predictions during training. While DETR performs attention operations in the spatial dimension, we apply it in the temporal dimension. This approach encourages our model to selectively aggregate features across frames. In the next section, we provide a detailed explanation of our method for extracting features and generating predictions from frame sequences.

\subsection{Model architecture}

\begin{figure*}[t]
    \centering
    \includegraphics[keepaspectratio,width=0.8\linewidth]{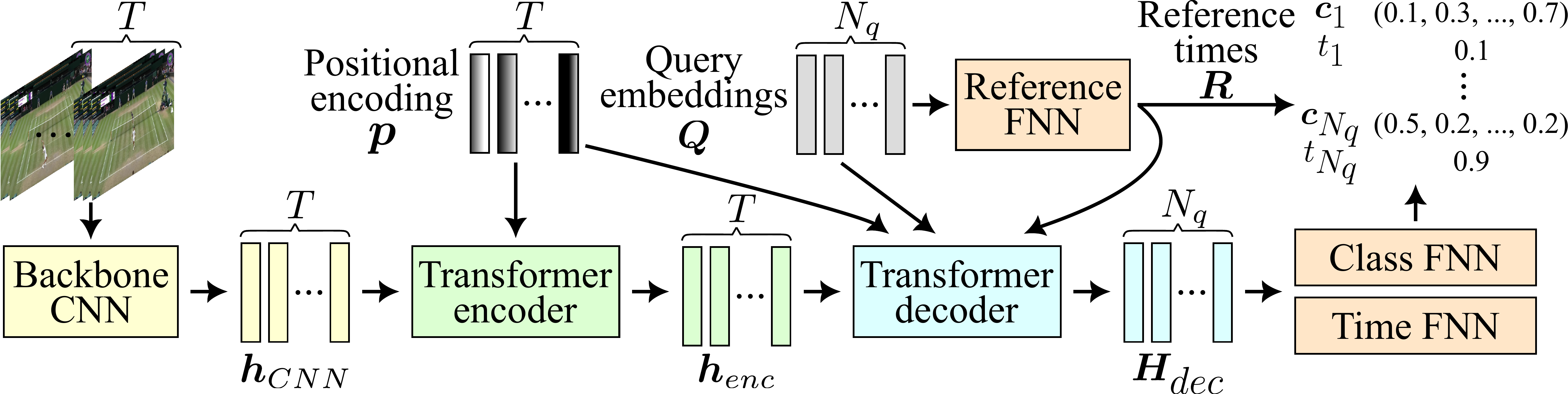}
    \caption{Model architecture of the proposed method.}
    \label{fig:arch}
    \vspace{-1.5em}
\end{figure*}

Figure~\ref{fig:arch} illustrates the architecture of our model. First, we convert an input frame sequence to a feature sequence. Given a frame sequence $\bm{x} \in \mathbb{R}^{3 \times T \times H \times W}$, a feature sequence $\bm{h}_{CNN} \in \mathbb{R}^{T \times D}$ is obtained with a backbone CNN $f_{CNN}(\cdot)$ as $\bm{h}_{CNN} = f_{CNN}(\bm{x})$, where $T$ is the length of the sequence, $H$ and $W$ are the height and width of the frames, and $D$ is the feature dimension. We employ small models of RegNet~\cite{ilija_cvpr2020} as our backbone CNN and incorporate gate-shift modules (GSM)~\cite{sudhakaran_cvpr2020} into the network. This design allows the network to extract features from lengthy sequences of frames with lightweight computation, enabling the network to capture differences between frames and effectively identify event frames.

A transformer encoder further enhances the extracted feature sequence. Given the extracted feature sequence $\bm{h}_{CNN}$, the enhanced feature sequence $\bm{h}_{enc} \in \mathbb{R}^{T \times D}$ is generated through $N_{enc}$ layers of a transformer encoder $f_{enc}(\cdot, \cdot)$ as $\bm{h}_{enc} = f_{enc}(\bm{h}_{CNN}, \bm{p})$, where $\bm{p} \in \mathbb{R}^{T \times D}$ represents a positional encoding. In line with the implementation of DETR~\cite{carion_eccv2020}, we add the positional encoding to the queries and keys of the self-attention modules rather than adding it to the feature sequence before inputting it into the encoder. This encoding process enables the features in the sequence to interact effectively through the self-attention modules, which aids in identifying event frames precisely.

The essential features are aggregated from the enhanced feature sequence $\bm{h}_{enc}$ using a set of $N_{q}$ query embeddings $\bm{Q} = \{\bm{q}_{i} \mid \bm{q}_{i} \in \mathbb{R}^{D} \}_{i=1}^{N_{q}}$ and $N_{dec}$ layers of a transformer decoder $f_{dec}(\cdot, \cdot, \cdot, \cdot)$. We adopt the decoding process from Conditional DETR~\cite{meng_iccv2021} rather than that from DETR~\cite{carion_eccv2020} to accelerate the convergence of training. In the decoding process, the query embeddings are transformed into reference frame times $\bm{R} = \{r_{i} \mid r_{i} \in \mathbb{R}\}_{i=1}^{N_{q}}$ using a feed-forward neural network (FNN) $f_{ref}(\cdot)$ as $r_{i} = f_{ref}(\bm{q}_{i})$. The reference frame times are utilized in the decoder to identify features in the feature sequence and are also employed in the final prediction process to present the base frame times. Given the positional encoding $\bm{p}$, the feature sequence $\bm{h}_{enc}$, the query embeddings $\bm{Q}$, and the reference times $\bm{R}$, feature embeddings $\bm{H}_{dec} = \{\bm{h}_{i}^{(dec)} \mid \bm{h}_{i}^{(dec)} \in \mathbb{R}^{D} \}_{i=1}^{N_{q}}$ are obtained as $\bm{H}_{dec} = f_{dec}(\bm{p}, \bm{h}_{enc}, \bm{Q}, \bm{R})$. The decoder enables our model to selectively aggregate event features from the feature sequence through cross-attention modules, resulting in precise event frame identification. While our primary goal is not to eliminate non-maximum suppression (NMS) in post-processing, the self-attention modules in the decoder facilitate the activation of one feature embedding for each event frame, similarly to how a single embedding is activated for each object in DETR~\cite{carion_eccv2020}. We validate the effectiveness of this learnable suppression in the temporal dimension in the supplementary material.

The feature embeddings are transformed into predictions through two FNNs: a class FNN and a time FNN. The class FNN is employed to obtain action class scores for events, while the time FNN is used to obtain the timing of these events. Given the feature embedding $\bm{h}_{i}^{(dec)}$, predictions are obtained as follows:
\begin{align}
    \hat{\bm{c}}_{i} ={} & f_{\sigma}\left(f_{class}\left(\bm{h}_{i}^{(dec)}\right)\right), \\
    \hat{t}_{i} ={} & f_{\sigma}\left(f_{time}\left(\bm{h}_{i}^{(dec)}\right) + r_{i}\right),
\end{align}
where $f_{class}(\cdot)$ and $f_{time}(\cdot)$ are the class and time FNNs, respectively, $f_{\sigma}(\cdot)$ is the sigmoid function, $\hat{\bm{c}}_{i} \in [0, 1]^{N_{c}}$ is a predicted class score vector with $N_{c}$ being the number of classes, and $\hat{t}_{i} \in [0, 1]$ is a predicted event frame time expressed as a normalized value in a sequence. Note that we use the sigmoid function instead of the softmax function to obtain the class scores because one benchmark dataset is structured in a multi-label format. In the subsequent section, we outline our proposed dynamic label assignment that dynamically assigns ground-truth labels to the predictions during training.

\subsection{Dynamic label assignment}

Instead of statically assigning ground-truth labels to predictions based on ground-truth and predicted times during training, we employ a dynamic approach that assigns the labels according to the predicted class scores and times. This dynamic assignment is essential for reducing the impact of temporal misalignment in labels because it can assign the labels to predictions that exhibit temporal offsets from the ground-truth times when the predicted scores align with the ground-truth action class labels. For this dynamic label assignment, we first calculate matching costs between ground-truth labels and predictions. Given a set of $N_{g}$ ground-truth labels $\bm{G} = \{(\bm{c}_{i}, t_{i}) \mid \bm{c}_{i} \in [0, 1]^{N_{c}}, t_{i} \in [0, 1]\}_{i=1}^{N_{g}}$, where $\bm{c}_{i}$ is a ground-truth action class label and $t_{i}$ is a ground-truth event frame time expressed as a normalized value in a sequence, the set is first padded with $\phi$, which indicates no event, to expand the set size to $N_{q}$ for one-to-one matching between ground truth labels and predictions. Once we have the padded ground-truth set, we proceed to calculate the matching cost $\mathcal{H}_{i, j}$ between the $i$-th ground truth label and the $j$-th prediction as follows:
\begin{align}
    \mathcal{H}_{i, j}^{(class)} ={} & -\frac{1}{N_{c}}\left[\bm{c}_{i}^{\intercal}\hat{\bm{c}}_{j} + \left(\bm{1} - \bm{c}_{i}\right)^{\intercal}\left(\bm{1} - \hat{\bm{c}}_{j}\right)\right], \label{eq:cost_class} \\
    \mathcal{H}_{i, j}^{(time)} ={} & \lvert t_{i} - \hat{t}_{j} \rvert, \\
    \mathcal{H}_{i, j} ={} & \mathbbm{1}_{\{i \not\in \bm{\Phi}\}}\left[\mathcal{H}_{i, j}^{(class)} + \lambda_{time}\mathcal{H}_{i, j}^{(time)}\right],
\end{align}
where $\bm{\Phi}$ is a set of indices corresponding to $\phi$, and $\lambda_{time}$ is a hyper-parameter that adjusts the influence of time costs during the matching process. The Hungarian algorithm~\cite{kuhn_naval1955} is applied to the calculated costs to identify the optimal assignment as $\hat{\omega} = \argmin_{\omega \in \bm{\Omega}_{N_{q}}}{\sum_{i=1}^{N_{q}}{\mathcal{H}_{i,\omega(i)}}}$, where $\bm{\Omega}_{N_{q}}$ is the set of all possible permutations of $N_{q}$ elements, and $\omega(i)$ is the $i$-th element of $\omega$. As indicated by Eq.~\ref{eq:cost_class}, the matching costs decrease when the predicted class scores align with ground-truth action class labels. This matching strategy allows predictions to exhibit temporal offsets from ground-truth frame times, particularly when neighboring frames are more likely to represent events, mitigating the impact of temporal misalignment in labels.

\subsection{Loss calculation}

The training loss is computed using the matched ground-truth labels and predictions. We employ focal loss~\cite{lin_iccv2017} for classification loss and L1 loss for frame time loss. The total loss $\mathcal{L}$ is calculated as follows:
\begin{align}
\mathcal{L}_{class} ={} & \frac{1}{N_{g}}\sum_{i=1}^{N_{q}}\left[\mathbbm{1}_{\left\{i \not\in \bm{\Phi}\right\}}\left(\bm{c}_{i} - \hat{\bm{c}}_{\hat{\omega}(i)}\right)^{2}f_{CE}\left(\bm{c}_{i}, \hat{\bm{c}}_{\hat{\omega}(i)}\right) + \right. \nonumber \\
&\quad\quad \left. \mathbbm{1}_{\left\{i \in \bm{\Phi}\right\}}\left(\bm{0} - \hat{\bm{c}}_{\hat{\omega}(i)}\right)^{2}f_{CE}\left(\bm{0}, \hat{\bm{c}}_{\hat{\omega}(i)}\right)\right], \label{eq:loss_class}\\
\mathcal{L}_{time} ={} & \frac{1}{N_{g}}\sum_{i=1}^{N_{q}}\mathbbm{1}_{\left\{i \not\in \bm{\Phi}\right\}} \lvert t_{i} - \hat{t}_{\hat{\omega}(i)} \rvert, \\
\mathcal{L} ={} & \mathcal{L}_{class} + \lambda_{time}\mathcal{L}_{time},
\end{align}
where $f_{CE}(\cdot, \cdot)$ is the binary cross-entropy loss function. Following Hong~\etal's work~\cite{hong_eccv2022}, we employ MixUp~\cite{zhang_iclr2018} and thus use soft labels for ground-truth class labels. To this end, we use the soft-target focal loss~\cite{dong_cvpr2020} as shown in Eq.~\ref{eq:loss_class}. We implement this focal loss approach for multi-label classification training on existing methods and re-evaluate their performances for fair comparisons in our experiments.

\subsection{Inference}

During the inference process, we need to convert the predictions into action class scores for each frame. To achieve this, we first transform the normalized frame times into unnormalized ones as $\hat{\tau}_{i} = f_{round}(\hat{t}_{i}T)$, where $f_{round}(\cdot)$ rounds the input value to the nearest integer, with a minimum value set to 1. Since multiple $\hat{\tau}_{\{1, \ldots, N_{q}\}}$ can point to a single frame, we summarize these results by applying the element-wise maximum function to the predicted score vectors $\hat{\bm{c}}_{i}$ and obtain the scores of the $i$-th frame $\bar{\bm{c}}_{i}$ as follows:
\begin{equation}
    \bar{\bm{c}}_{i} = \begin{cases}
        \max(\hat{\bm{\mathcal{C}}}_{i}) & \text{if $|\hat{\bm{\mathcal{C}}}_{i}| > 0$}, \\
        \bm{0} & \text{otherwise},
    \end{cases}
\end{equation}
where $\hat{\bm{\mathcal{C}}}_{i} = \{\hat{\bm{c}}_{j} \mid \hat{\tau}_{j} = i\}$ is the set of the predicted class score vectors whose paired frame times $\hat{\tau}_{j}$ are equal to $i$, and $\max(\cdot)$ outputs a score vector generated by taking the element-wise maximum values of the vectors in the set. Finally, the frame-wise action class scores are obtained as $\bar{\bm{C}} = \{(i, \bar{\bm{c}}_{i})\}_{i=1}^{T}$. 
\section{Experiments}
\label{sec:experiments}

To demonstrate the effectiveness of our method, we conduct experiments using two different sets of labels. The first set consists of the precise labels provided by Hong~\etal~\cite{hong_eccv2022}. The experimental results with these labels show how our method mitigates the effect of the unavoidable temporal misalignment caused by human errors or unclear event boundaries. The second set contains noisy labels generated by adding Gaussian noise to the event times in the precise labels. In the experiments with these labels, we assume that the event times are approximately annotated and analyze how our method performs with these noisy labels.

We use Spot~\cite{hong_eccv2022} and T-DEED~\cite{xarles_cvprw2024} as baseline methods. Spot and T-DEED are state-of-the-art precise action-spotting methods that demonstrate substantial performance on the benchmark datasets. We verify the effectiveness of our method by comparing the results.

\subsection{Datasets}
\label{subsec:dataset}


We conduct experiments with four publicly available benchmark datasets: Tennis~\cite{zhang_tog2021}, FineDiving~\cite{xu_cvpr2022}, Figure Skating (FS)~\cite{hong_iccv2021}\footnote{Due to copyright restrictions, we were unable to download one video in the FS dataset. All the methods in our experiments are evaluated without the video.}, and FineGym (FG)~\cite{shao_cvpr2020}. The videos in these datasets are divided into clips. The labels include event times indicated by the frame indices within each clip, along with the action classes performed in those frames. The data are split into training, validation, and test sets. Unless otherwise stated, we report performance on the test sets using model parameters that yield the best results on the validation sets. We observe that multiple action labels are tagged to certain frames in the FG dataset and thus treat the dataset as a multi-label dataset.


Following Hong~\etal's work~\cite{hong_eccv2022}, evaluations are conducted using two different configurations for the FS and FG datasets. For the FS dataset, the configurations are FS-Comp and FS-Perf. For the FG dataset, the configurations are FG-Full and FG-Start. Please refer to the supplementary material for more details.


We create four types of noisy labels by adding Gaussian noise with four different standard deviations: $\sigma = 0.5$, $1.0$, $1.5$, and $2.0$. We then separately evaluate the performance using the labels associated with each deviation. As the standard deviation increases, the event times deviate more from their originally tagged frame times, which complicates the learning of consistent event frame identification. Note that Gaussian noise is only added to the labels in the training sets. We assume that there are no precise labels for validation and thus report the performance on the test sets using the model parameters from the final epoch in our experiments with the noisy labels.

\subsection{Evaluation metrics}

We use mean average precision (mAP) as our evaluation metric. During the calculation of AP, we introduce a tolerance level, denoted as $\delta$. This tolerance allows a prediction to be counted as a true positive even if it is $\delta$ frames away from the corresponding ground-truth frames.

\subsection{Implementaiton details}
\label{subsec:imple}

To ensure fair comparisons, we follow existing works~\cite{hong_eccv2022,xarles_cvprw2024} by using the same backbone CNN, input pre-processing, and output post-processing methods. Specifically, we utilize RegNetY 200MF and 800MF~\cite{ilija_cvpr2020} with GSM~\cite{sudhakaran_cvpr2020} as our backbone CNN initialized with the ImageNet~\cite{deng_cvpr2009} pre-trained model. The frame sequence length $T$ is set to 100. We resize the frames to either 224 or 256 pixels in height before cropping them to a size of $224 \times 224$. During training, we employ standard data augmentation techniques, followed by MixUp~\cite{zhang_iclr2018}. Soft NMS~\cite{bodla_iccv2017} is applied to the outputs of our model. Note that soft NMS is applied to the outputs of all the methods in our experiments.

The hyper-parameter $\lambda_{time}$ is set to 10 unless otherwise specified. For additional details regarding the implementation, please refer to the supplementary material.

\subsection{Evaluation with precise labels}
\label{subsec:eval_precise}

\subsubsection{Comparison against state-of-the-art methods}
\label{subsubsec:comp_precise_sota}

\begin{table*}[t]
    \caption{Comparison against state-of-the-art methods. The models are trained with precise labels.}
    \label{table:comp_precise}
    \centering
    \setlength{\tabcolsep}{5.3pt}
    \small
    \begin{tabular}{@{}llcccccccccccc@{}}
        \toprule
        && \multicolumn{2}{c}{Tennis} & \multicolumn{2}{c}{FineDiving} & \multicolumn{2}{c}{FS-Comp} & \multicolumn{2}{c}{FS-Perf} & \multicolumn{2}{c}{FG-Full} & \multicolumn{2}{c}{FG-Start} \\
        \cmidrule(lr){3-4}\cmidrule(lr){5-6}\cmidrule(lr){7-8}\cmidrule(lr){9-10}\cmidrule(lr){11-12}\cmidrule(lr){13-14}
        Backbone & Method & $\delta=1$ & $\delta=2$ & $\delta=1$ & $\delta=2$ & $\delta=1$ & $\delta=2$ & $\delta=1$ & $\delta=2$ & $\delta=1$ & $\delta=2$ & $\delta=1$ & $\delta=2$ \\
        \midrule
        \multirow{3}{*}{\shortstack[c]{RegNetY\\ 200MF}} & Spot~\cite{hong_eccv2022} & 95.63 & 97.83 & 70.47 & 88.03 & 77.24 & 89.41 & 89.22 & 96.59 & 55.20 & 67.18 & 68.22 & 80.92 \\
        & T-DEED~\cite{xarles_cvprw2024} & 95.29 & 98.05 & \textbf{75.44} & 90.08 & 77.71 & 86.32 & 90.82 & 96.63 & 56.34 & 68.09 & 69.29 & 81.50 \\
        & Ours & 96.21 & \textbf{98.15} & 70.69 & 87.88 & 80.93 & 93.64 & 87.82 & 97.56 & 55.87 & 70.26 & 67.90 & 82.04 \\
        \multicolumn{14}{@{}c@{}}{\makebox[\linewidth]{\dashrule[black]}} \\
        \multirow{4}{*}{\shortstack[c]{RegNetY\\ 800MF}} & Spot~\cite{hong_eccv2022} & 96.16 & 97.95 & 66.13 & 85.76 & 84.82 & 93.39 & 90.8 & 97.30 & 56.27 & 69.32 & 69.22 & 82.64 \\
        & Spot (+Flow)~\cite{hong_eccv2022} & 96.25 & 98.11 & 70.12 & 87.15 & \textbf{86.65} & \textbf{94.67} & \textbf{91.70} & \textbf{97.78} & \textbf{57.53} & 69.61 & \textbf{70.58} & 83.00 \\
        & T-DEED~\cite{xarles_cvprw2024} & 96.11 & 98.01 & 75.14 & 88.94 & 72.58 & 81.54 & 91.04 & 97.54 & 57.19 & 69.95 & 70.10 & 82.66 \\
        & Ours & \textbf{96.61} & 98.07 & 72.42 & \textbf{90.14} & 80.55 & 91.07 & 87.54 & 96.95 & 56.55 & \textbf{71.05} & 68.76 & \textbf{83.36} \\
        \bottomrule
    \end{tabular}
    \vspace{-0.5em}
\end{table*}

Table~\ref{table:comp_precise} illustrates the comparison results. As shown in the table, our method demonstrates competitive or even superior performance on the Tennis, FineDiving, and FG datasets. Remarkably, our method does not utilize the enhanced gate-shift modules~\cite{sudhakaran_tpami2023} used in T-DEED, nor does it incorporate the flow modality used in Spot. Despite this, our method achieves significant performance gains. These competitive results can be attributed to our strategy of effectively addressing the temporal misalignment in the labels through our dynamic label assignment.

\begin{figure*}[t]
    \centering
    \begin{subfigure}{0.32\linewidth}
        \includegraphics[keepaspectratio,width=\linewidth]{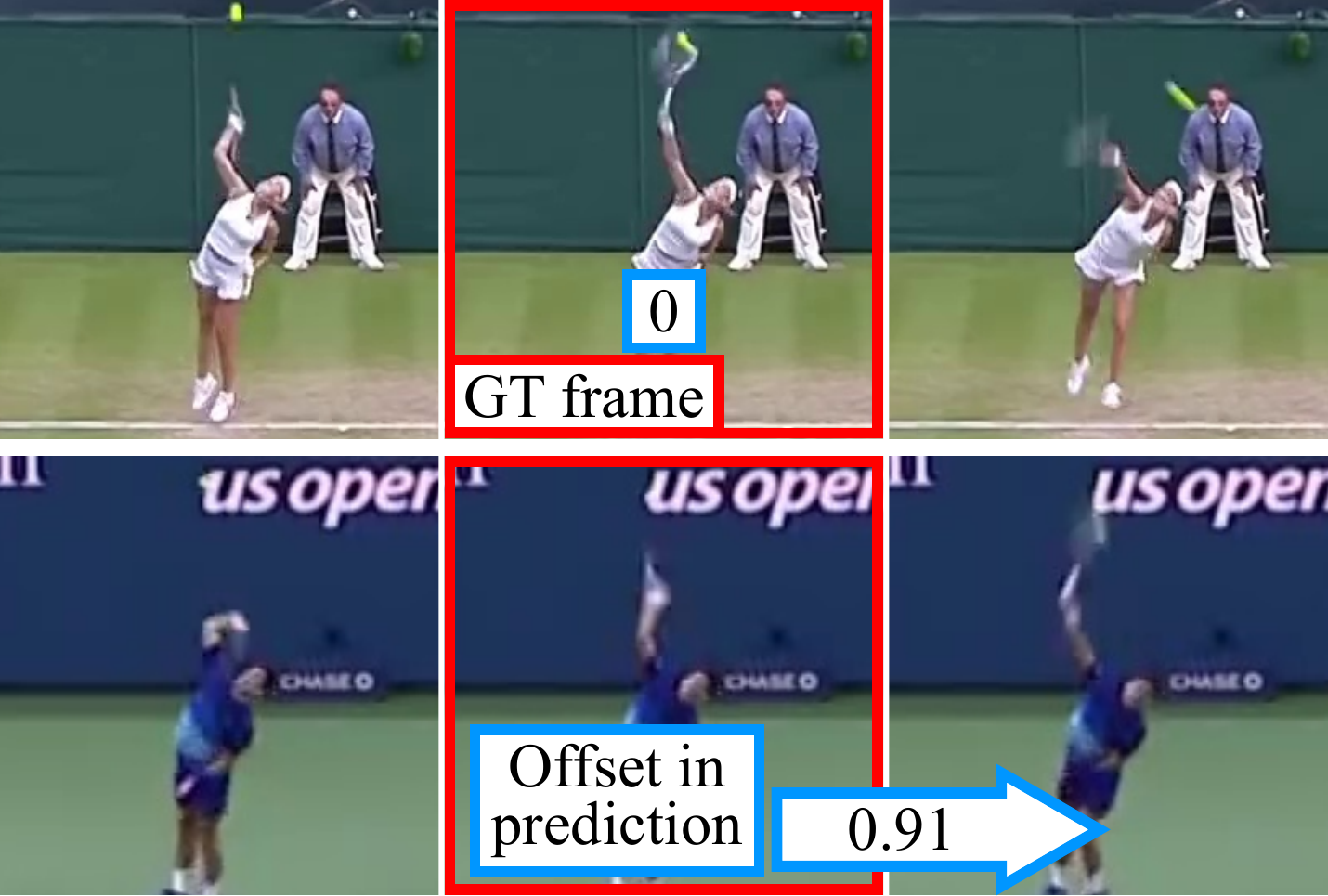}
        \caption{Far court serve}
        \label{fig:offset_far_court_serve}
    \end{subfigure}
    \hfill
    \begin{subfigure}{0.32\linewidth}
        \includegraphics[keepaspectratio,width=\linewidth]{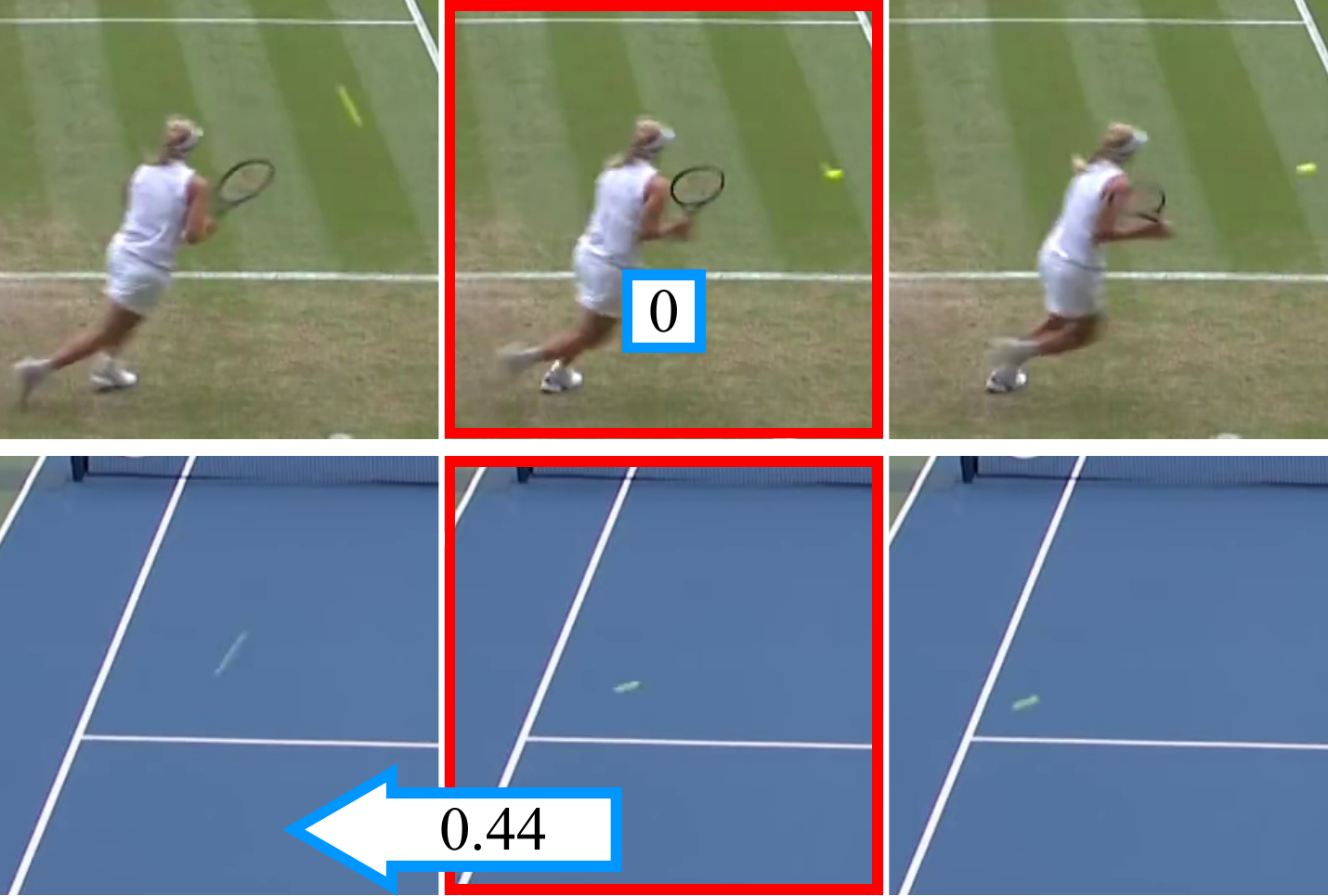}
        \caption{Near court bounce}
        \label{fig:offset_near_court_bounce}
    \end{subfigure}
    \hfill
    \begin{subfigure}{0.32\linewidth}
        \includegraphics[keepaspectratio,width=\linewidth]{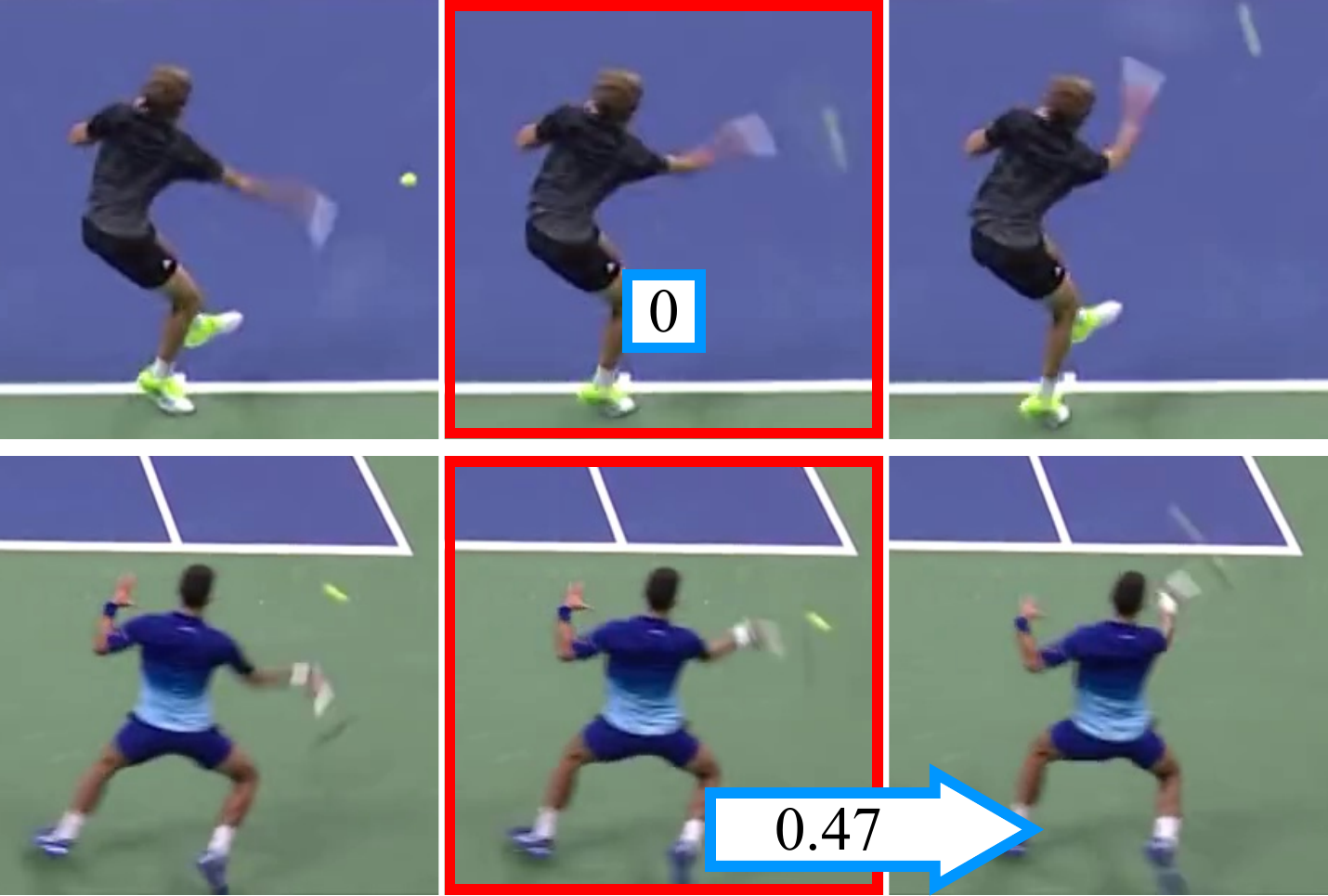}
        \caption{Near court swing}
        \label{fig:offset_near_court_swing}
    \end{subfigure}
    \caption{Example cases from the Tennis dataset, where the predicted times exhibit offsets from the ground-truth times during training. The center images display cropped action regions from the tagged frames, while the left and right images show the regions from the frames immediately before and after the tagged frames. The values indicated in the blue boxes and arrows represent the offsets in the predictions.}
    \label{fig:offset}
    \vspace{-1.5em}
\end{figure*}

Figure~\ref{fig:offset} illustrates example cases from the Tennis dataset, which show that predicted times in predictions that match ground-truth labels may have temporal offsets during training. The central image in each case shows the cropped action region of the tagged frame, while the left and right images capture the regions of the frames immediately before and after the tagged frames. The values displayed in the blue boxes and arrows indicate the offsets in the predictions relative to the tagged frames. In the upper images of each action, where the actions appear to occur within the tagged frames, the ground-truth labels are assigned to the predictions that do not exhibit offsets. In contrast, when the actions seem to take place in the adjacent frames or between the frames, the ground-truth labels are assigned to the predictions having reasonable offsets, as demonstrated in the lower images of each action. These results indicate that predictions can deviate reasonably from ground-truth times when necessary due to our dynamic label assignment, resulting in improved training consistency compared to static label assignment to frames.

Our method demonstrates lower performance compared to Spot on the FS dataset. This reduced performance is likely due to the limited size of the dataset, which contains fewer than ten videos and only several hundred samples for each action class. This quantity is significantly smaller than those of the other datasets. Since we use transformers, our model requires a substantial number of samples for training and thus may show limited performance with small datasets. This assumption is further supported by our results with the 200MF and 800MF backbones, which indicate no performance improvement with the larger backbone.

\subsubsection{Component analysis}
\label{subsubsec:component}

\begin{table*}[t]
    \caption{Comparison with various configurations of our method. The models of all the configurations are trained with precise labels.}
    \label{table:comp_design}
    \centering
    \setlength{\tabcolsep}{4.8pt}
    \small
    \begin{tabular}{@{}ccccccccccccccccc@{}}
        \toprule
        &\multicolumn{2}{c}{Design} & \multicolumn{2}{c}{Matching} & \multicolumn{2}{c}{Tennis} & \multicolumn{2}{c}{FineDiving} & \multicolumn{2}{c}{FS-Comp} & \multicolumn{2}{c}{FS-Perf} & \multicolumn{2}{c}{FG-Full} & \multicolumn{2}{c}{FG-Start} \\
        \cmidrule(lr){2-3}\cmidrule(lr){4-5}\cmidrule(lr){6-7}\cmidrule(lr){8-9}\cmidrule(lr){10-11}\cmidrule(lr){12-13}\cmidrule(lr){14-15}\cmidrule(lr){16-17}
        &Enc. & Dec. & Time & Class & $\delta=1$ & $\delta=2$ & $\delta=1$ & $\delta=2$ & $\delta=1$ & $\delta=2$ & $\delta=1$ & $\delta=2$ & $\delta=1$ & $\delta=2$ & $\delta=1$ & $\delta=2$ \\
        \midrule
        (c1) & \checkmark && \checkmark && 95.84 & 97.8 & 61.41 & 82.79 & 80.35 & 90.38 & \textbf{89.26} & \textbf{97.32} & \textbf{57.35} & 70.29 & 69.3 & 81.84 \\
        (c2) & \checkmark & \checkmark & \checkmark && 96.23 & 97.9 & 71.7 & 88.62 & 77.96 & 90.53 & 86.52 & 96.45 & 57.22 & \textbf{71.49} & \textbf{69.54} & \textbf{83.86} \\
        (c3) & \checkmark & \checkmark & \checkmark & \checkmark & \textbf{96.61} & \textbf{98.07} & \textbf{72.42} & \textbf{90.14} & \textbf{80.55} & \textbf{91.07} & 87.54 & 96.95 & 56.55 & 71.05 & 68.76 & 83.36 \\
        \bottomrule
    \end{tabular}
    \vspace{-0.5em}
\end{table*}

To assess the effectiveness of the components in our method, we evaluate our model using three distinct configurations and compare the results. The comparison results are presented in Table~\ref{table:comp_design}. In configuration (c1), predictions are generated from frame features obtained from the encoder, with ground-truth labels statically assigned to predictions derived from tagged frames. (c2) generates predictions based on feature embeddings from the decoder, allowing predicted times to deviate from the ground-truth times. However, action class scores are not used in the matching process, encouraging the assignment of ground-truth labels to predictions that accurately identify ground-truth times. (c3) is our proposed dynamic label assignment method. The results indicate that allowing offsets in predictions, as seen in (c2), enhances performance, and incorporating class scores during the matching process in (c3) further improves performance for the Tennis and FineDiving datasets. These results indicate the effectiveness of the proposed dynamic label assignment across these datasets.

The proposed dynamic label assignment strategy, implemented in (c3), does not improve performance on the FS and FG datasets. This lack of enhancement can be attributed to the ambiguity inherent in the definitions of event frames within these datasets. It seems that frames adjacent to tagged frames may also be considered event frames due to this ambiguity. We discuss this issue further in Sec~\ref{subsubsec:loc_weight_analysis}.

\subsection{Evaluation with noisy labels}
\label{subsec:eval_noise}

In our experiments with noisy labels, RegNetY 200MF serves as the backbone for all the models. For the Tennis and FineDiving datasets, we set $\lambda_{time}$ to 8, 4, 2, and 1 for the labels associated with $\sigma$ of 0.5, 1.0, 1.5, and 2.0, respectively. Conversely, for the FS and FG datasets, $\lambda_{time}$ is set to 8 for the labels with all the $\sigma$ values. This variation is attributed to the event frame ambiguity mentioned in Sec.~\ref{subsubsec:component}, which we analyze in this section. We report the mean and standard deviation of mAP values obtained from 10 trials using 10 different random seeds because we noted significant fluctuations in mAP values between the trials.

\subsubsection{Comparison against state-of-the-art methods}

\begin{figure*}[t]
    \centering
    \begin{subfigure}{0.3\linewidth}
        \includegraphics[keepaspectratio,width=\linewidth]{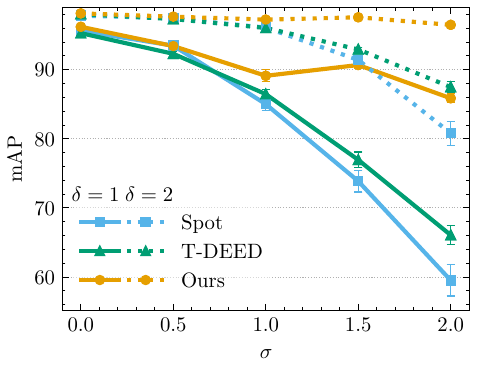}
        \caption{Tennis}
        \label{fig:noise_tennis}
    \end{subfigure}
    \begin{subfigure}{0.3\linewidth}
        \includegraphics[keepaspectratio,width=\linewidth]{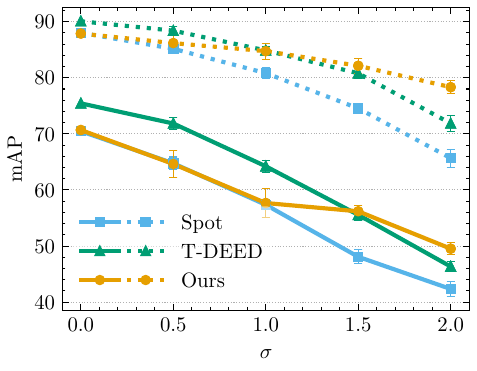}
        \caption{FineDiving}
        \label{fig:finediving}
    \end{subfigure}
    \begin{subfigure}{0.3\linewidth}
        \includegraphics[keepaspectratio,width=\linewidth]{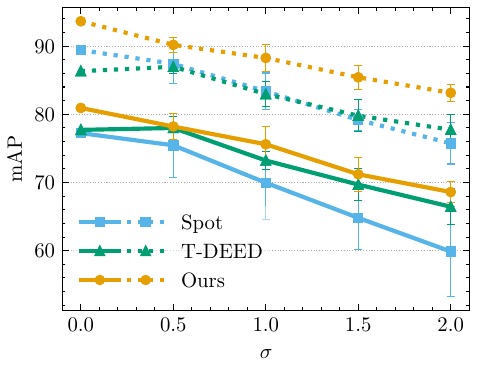}
        \caption{FS-Comp}
        \label{fig:noise_fscomp}
    \end{subfigure} \\
    \begin{subfigure}{0.3\linewidth}
        \includegraphics[keepaspectratio,width=\linewidth]{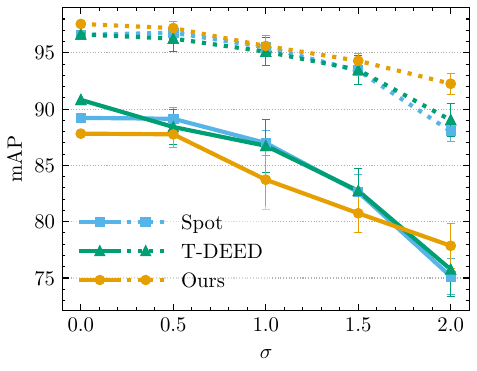}
        \caption{FS-Perf}
        \label{fig:noise_fsperf}
    \end{subfigure}
    \begin{subfigure}{0.3\linewidth}
        \includegraphics[keepaspectratio,width=\linewidth]{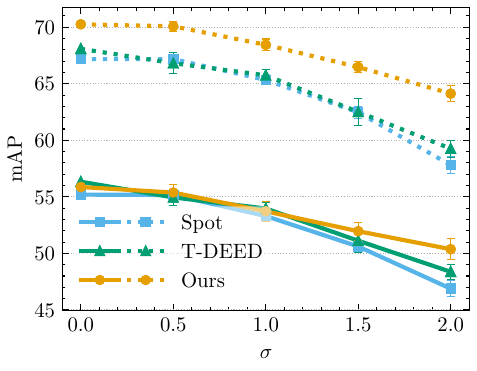}
        \caption{FG-Full}
        \label{fig:noise_fgfull}
    \end{subfigure}
    \begin{subfigure}{0.3\linewidth}
        \includegraphics[keepaspectratio,width=\linewidth]{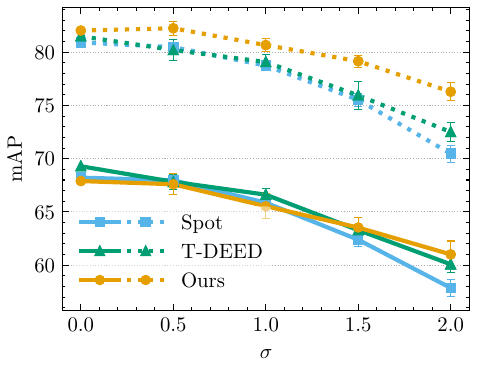}
        \caption{FG-Start}
        \label{fig:noise_fgstart}
    \end{subfigure}
    \caption{Comparison against state-of-the-art methods. The solid and dotted lines indicate the results for $\delta = 1$ and $\delta = 2$, respectively. For reference, we include the results from Table~\ref{table:comp_precise} at $\sigma$ of 0.0.}
    \label{fig:noise}
    \vspace{-1.5em}
\end{figure*}

Figure~\ref{fig:noise} shows the mAP values in relation to the standard deviations of the added Gaussian noise. The solid and dotted lines indicate results for $\delta = 1$ and $2$, respectively. For reference, we also include results from Table~\ref{table:comp_precise} at $\sigma = 0.0$. The figure demonstrates that our method surpasses existing methods when considerable noise is introduced into the training labels. In particular, our method achieves 19.79 and 9.00 higher mAP for $\delta = 1$ and $2$, respectively, when $\sigma = 2.0$, compared to the existing methods on the Tennis dataset. The event frames in the Tennis dataset are visually distinct compared to those in the other datasets. This clarity is vital for our method, as our dynamic label assignment relies on predicted class scores to correctly assign ground-truth labels to predictions. When event frames are clear, predicted class scores become more distinctive, enabling our method to accurately assign ground-truth labels to predictions even if they deviate significantly from the times of the ground-truth labels. This is the reason why we employ lower $\lambda_{time}$ values for higher $\sigma$ values to rely more on predicted scores during the matching process for the Tennis and FineDiving datasets, where event frames are more apparent. In contrast, we maintain $\lambda_{time} = 8$ across all $\sigma$ values for the FS and FG datasets, where event frames are less distinct.

\begin{figure}[t]
    \centering
    \includegraphics[keepaspectratio,width=0.7\linewidth]{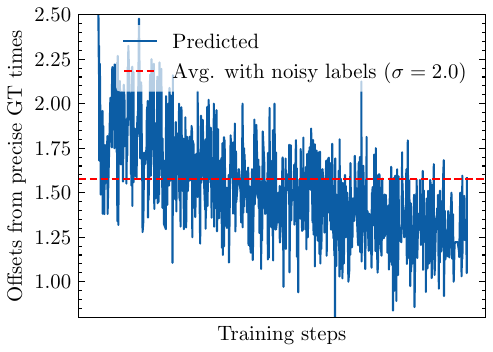}
    \caption{Offsets between ground-truth times in the original precise labels and predicted times. Since our model is trained using the noisy labels associated with $\sigma$ of 2.0, we can expect an average offset of 1.58 between the predicted times and the ground truth times in the precise labels. However, the predicted times eventually exhibit offsets of less than 1.58 during training.}
    \label{fig:noise_reduct}
    \vspace{-1.5em}
\end{figure}

Since precise labels are provided, we can verify whether our method effectively cancels the Gaussian noise injected into the noisy labels. Figure~\ref{fig:noise_reduct} shows the temporal offsets in predictions relative to the ground-truth times in the precise labels during training with the Tennis dataset. Since the model is trained using noisy labels that incorporate Gaussian noise with a standard deviation of $\sigma = 2.0$, we can expect an average offset time of 1.58 in the predicted times relative to the ground-truth times in the precise labels. However, as training progresses, the predicted times exhibit an offset time of less than 1.58. This outcome suggests that our dynamic label assignment does not strictly adhere to the times in the noisy labels. Instead, it relies on action class scores to identify event frames, enabling our model to learn to identify event frames consistently, even when event times are approximately annotated.

\begin{figure*}[t]
    \centering
    \hfill
    \begin{minipage}{0.61\linewidth}
        \begin{subfigure}{0.492\linewidth}
            \includegraphics[keepaspectratio,width=\linewidth]{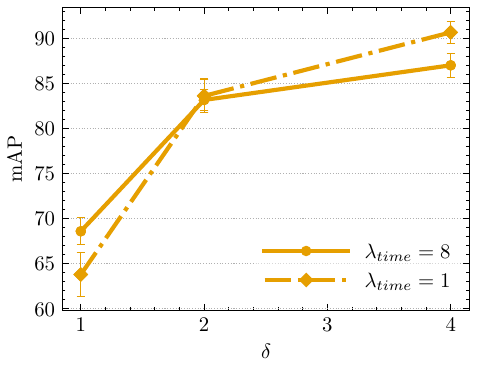}
            \caption{FS-Comp}
            \label{fig:noise_offset_fscomp}
        \end{subfigure}
        \hfill
        \begin{subfigure}{0.492\linewidth}
            \includegraphics[keepaspectratio,width=\linewidth]{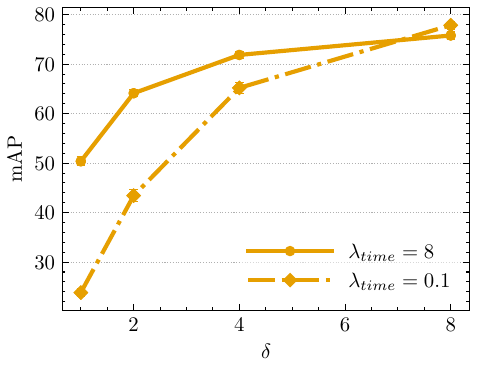}
            \caption{FG-Full}
            \label{fig:noise_offset_fgfull}
        \end{subfigure}
        \caption{Comparison with different $\lambda_{time}$ values. The noisy labels associated with $\sigma = 2.0$ are used for training.}
        \label{fig:noise_offset}
    \end{minipage}
    \hfill
    \begin{minipage}{0.25\linewidth}
        \centering
        \includegraphics[keepaspectratio,width=\linewidth]{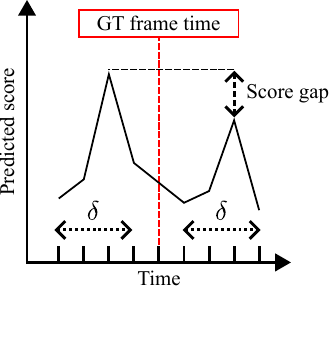}
        \caption{Calculated score gaps.}
        \label{fig:score_gap}
    \end{minipage}
    \hfill
    \vspace{-0.5em}
\end{figure*}

\begin{figure*}[t]
    \centering
    \includegraphics[keepaspectratio,width=0.75\linewidth]{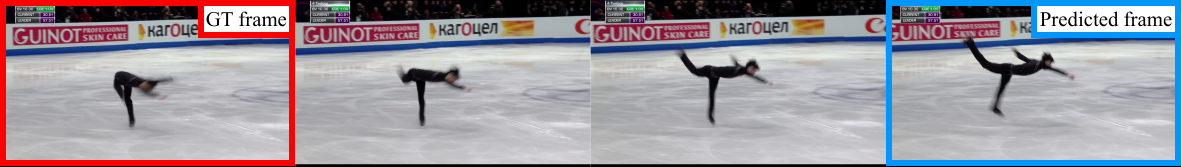}
    \caption{Example case where our model recognizes the action class ``Spin take off" correctly but identifies a frame apart from the tagged frame as an event frame in the test set of the FS dataset.}
    \label{fig:offset_weight}
    \vspace{-1.4em}
\end{figure*}

\subsubsection{Time weight analysis}
\label{subsubsec:loc_weight_analysis}

In this section, we adjust $\lambda_{time}$ to 1 and 0.1 for the FS and FG datasets, respectively, in training with the noisy labels generated by Gaussian noise of $\sigma = 2.0$ and evaluate the performance under these parameters.

Figure~\ref{fig:noise_offset} presents a comparison of our model's performance with the high and low values of $\lambda_{time}$. As shown in the figure, our model performs better with the higher $\lambda_{time}$ values when the tolerance level $\delta$ is low. However, as the tolerance increases, the model trained with the lower $\lambda_{time}$ values demonstrates superior performance. This suggests that while our model is more effective at recognizing events with low $\lambda_{time}$ values, it also tends to identify frames that are apart from tagged frames as event frames.

Figure~\ref{fig:offset_weight} shows an example where our model accurately recognizes the action ``Spin take off" but incorrectly identifies a frame apart from the tagged frame as an event frame in the test set of the FS dataset. The figure illustrates that the label is assigned to a frame where it is ambiguous whether the player's leg is off the ground, while our model identifies the frame where the player clearly takes off. We frequently encounter these incorrectly identified frames. Since our dynamic label assignment is based on predicted class scores, our method tends to assign ground-truth labels to predictions that point to frames where actions are distinctly observable. Consequently, our model identifies frames subsequent to tagged frames as event frames in the FS dataset. More examples are provided in the supplementary material.

\begin{table}[t]
    \caption{Average score gaps between the highest and the second highest scores.}
    \label{table:score_ow}
    \centering
    \small
    \begin{tabular}{@{}ccccc@{}}
        \toprule
        & FS-Comp & FS-Perf & FG-Full & FG-Start \\
        $\lambda_{time}$ & ($\delta=4$) & ($\delta=4$) & ($\delta=8$) & ($\delta=8$) \\
        \midrule
        8 & 0.08 & 0.08 & 0.04 & 0.05 \\
        1 (FS) / 0.1 (FG) & 0.27 & 0.27 & 0.06 & 0.07 \\
        \bottomrule
    \end{tabular}
    \vspace{-1.5em}
\end{table}

We further analyze the reason behind the improved performance with low $\lambda_{time}$ values by examining the predicted class scores around ground-truth times in the test sets. We calculate the score gaps between the highest and second-highest predicted scores corresponding to ground-truth classes within a tolerance $\delta$ from the ground-truth times, as illustrated in Fig.~\ref{fig:score_gap}. We focus on samples that have no other ground-truth labels around target labels. Since only one event frame is present within the tolerance, our model is expected to produce one high score within the tolerance, leading to large score gaps. Table~\ref{table:score_ow} shows the average score gaps for different $\lambda_{time}$ values. As seen in the table, the score gaps increase as $\lambda_{time}$ values decrease. These results indicate that our model identifies a single event frame with greater confidence when trained with low $\lambda_{time}$ values. When $\lambda_{time}$ is set high, our dynamic label assignment needs to adhere strictly to ground-truth times during training, even when the action appearances in tagged frames are visually inconsistent between samples. Conversely, with low $\lambda_{time}$, our assignment can rely more on the predicted scores, allowing our assignment strategy to select event frames that are visually consistent. Consequently, the model trained with lower $\lambda_{time}$ values can select event frames confidently and thus achieve better performance.
\section{Conclusion}
\label{sec:conclusion}
We propose a dynamic label assignment strategy aimed at improving precise action-spotting performance by tackling the issue of temporal misalignment in ground-truth labels. This method assigns labels to predictions based not only on the predicted times but also on the predicted class scores. As a result, the predicted times can deviate from the times in the labels, effectively addressing the misalignment. Experiments show that our method outperforms existing methods, particularly when events are visually clear, and the labels are significantly misaligned with the actual event times.
{
    \small
    \bibliographystyle{ieeenat_fullname}
    \bibliography{main}
}

\clearpage
\setcounter{page}{1}
\maketitlesupplementary

\setcounter{section}{0}
\renewcommand*{\thesection}{\Alph{section}}

\setcounter{figure}{0}
\renewcommand*{\thefigure}{\roman{figure}}

\setcounter{table}{0}
\renewcommand*{\thetable}{\roman{table}}

\section{Dataset details}

As detailed in Sec.~{4.1} of the main paper, evaluations are conducted using two different configurations for the FS~\cite{hong_iccv2021} and FG~\cite{shao_cvpr2020} datasets.

For the FS dataset, there are two types of splits: a competition split (FS-Comp), which withholds certain videos from the training set for testing, and a performance split (FS-Perf), which divides clips from each video into training, validation, and test sets. The FS-Comp split is utilized to evaluate the generalization capabilities of models, as the withheld videos contain different views and backgrounds from those in the training set. In contrast, the FS-Perf split assesses spotting performance when certain domain information, such as views and backgrounds, is available in the training data. 

For the FG dataset, we present results for two scenarios: spotting both the start and end of actions in the full setting (FG-Full) and spotting only the start of the actions in the start setting (FG-Start). Hong~\etal~\cite{hong_eccv2022} noted that the annotations for action start frames are generally more visually consistent than those for end frames. Consequently, the performance for the FG-Start split is reported separately.

\section{Further implementation details}

We set both the number of encoder layers $N_{enc}$ and decoder layers $N_{dec}$ to 6, and configure the number of queries $N_{q}$ to 100. In the multi-head attention mechanism of the encoder and decoder layers, the number of heads is set to 8. The ReLU activation function is used in the FNNs of the encoder and decoder layers. The feature dimension is expanded to 2048 when the features pass through the activation functions. For positional encoding, we use a typical 1D sinusoidal positional encoding. The feature dimension $D$ is set to 368 for the RegNetY 200MF backbone CNN and 768 for the RegNetY 800MF backbone CNN~\cite{ilija_cvpr2020}. We observed unstable training on the FS and FG datasets with the RegNetY 800MF backbone CNN. To address this, we employed a reparameterization technique~\cite{zhai_icml2023} to stabilize the training process.

The reference, class, and time FNNs are built using the linear layers and the ReLU activation function. The reference FNN consists of 2 linear layers with the activation function applied between them. The class FNN contains a single linear layer. The time FNN is made up of 3 linear layers, each separated by the activation function. Throughout this process, the feature dimension remains constant at $D$ even when it passes through the activation functions.

To train our model, we use the AdamW~\cite{loshchiloy_iclr2019} optimizer with a batch size of 8, whose initial learning rates for the backbone CNN and the transformer are set to $10^{-3}$ and $10^{-4}$, respectively. Weight decay is set to $10^{-4}$. Our model is trained for 200 epochs on the FG dataset and for 100 epochs on the other datasets. The learning rates are adjusted with a linear warmup for the initial 3 epochs followed by the cosine decay~\cite{loshchilov_iclr2017}. To enhance performance, we incorporate auxiliary losses based on predictions from all the decoder layers, following the training method of DETR~\cite{carion_eccv2020}.

As detailed in Sec.~{4.3} of the main paper, we follow existing methods~\cite{hong_eccv2022,xarles_cvprw2024} by implementing both input pre-processing and output post-processing techniques to ensure fair comparisons. We generate batches by randomly sampling $T$-length clips from the training videos. Each training epoch is formed by grouping 625 training steps, amounting to 500,000 frames per epoch. For data augmentation, we employ random cropping, random horizontal flipping, color jittering, Gaussian blur, and MixUp~\cite{zhang_iclr2018}. In the FS dataset, we apply the label dilation technique, which extends event labels to one frame before and one frame after each event during training to alleviate label sparsity. During testing, we input frame sequences with a \SI{50}{\percent} overlap and average the predictions for each frame. We extract predictions with scores greater than 0.01 to calculate mAP values. The window size for soft NMS is set to 3.

\section{Comparison without soft NMS}

\begin{table*}[t]
    \caption{Comparison against state-of-the-art methods without soft NMS. The models are trained with precise labels.}
    \label{table:comp_precise_nonms}
    \centering
    \begin{tabular}{@{}lcccccccccccc@{}}
        \toprule
        & \multicolumn{2}{c}{Tennis} & \multicolumn{2}{c}{FineDiving} & \multicolumn{2}{c}{FS-Comp} & \multicolumn{2}{c}{FS-Perf} & \multicolumn{2}{c}{FG-Full} & \multicolumn{2}{c}{FG-Start} \\
        \cmidrule(lr){2-3}\cmidrule(lr){4-5}\cmidrule(lr){6-7}\cmidrule(lr){8-9}\cmidrule(lr){10-11}\cmidrule(lr){12-13}
        Method & $\delta=1$ & $\delta=2$ & $\delta=1$ & $\delta=2$ & $\delta=1$ & $\delta=2$ & $\delta=1$ & $\delta=2$ & $\delta=1$ & $\delta=2$ & $\delta=1$ & $\delta=2$ \\
        \midrule
        Spot~\cite{hong_eccv2022} & 96.02 & 97.32 & 65.68 & 83.96 & 73.91 & 82.48 & \textbf{82.43} & \textbf{89.58} & 41.87 & 47.04 & 53.82 & 59.72 \\
        Spot (+Flow)~\cite{hong_eccv2022} & 95.12 & 96.41 & 67.72 & 82.92 & 73.33 & 80.35 & 81.04 & 86.48 & 41.48 & 46.02 & 53.84 & 59.13 \\
        T-DEED~\cite{xarles_cvprw2024} & 95.59 & 96.99 & 62.75 & 74.94 & 21.5 & 23.56 & 81.03 & 86.29 & 40.12 & 44.78 & 51.16 & 56.71 \\
        Ours & \textbf{96.56} & \textbf{97.74} & \textbf{72.60} & \textbf{89.13} & \textbf{74.23} & \textbf{84.07} & 80.50 & 88.64 & \textbf{46.85} & \textbf{54.43} & \textbf{59.22} & \textbf{68.05} \\
        \bottomrule
    \end{tabular}
\end{table*}

Although eliminating NMS is not our target, the proposed method, by its nature, enables our model to learn to suppress duplicated prediction results with the self-attention modules in the decoder. To confirm the effectiveness, we compare results without soft NMS~\cite{bodla_iccv2017}. Table~\ref{table:comp_precise_nonms} illustrates the comparison results with the 800MF backbone CNN. Note that the label dilation used in Spot~\cite{hong_eccv2022} and our method and the label displacement used in T-DEED~\cite{xarles_cvprw2024} are disabled so that the models output one prediction result for each event. As shown in the table, our method outperforms the existing methods on almost all the datasets. In particular, our method achieves 4.88, 4.98, and 5.38 higher mAP on the FineDiving, FG-Full, and FG-Start datasets with $\delta=1$. These results indicate that the suppression with the self-attention modules in the decoder performs better than that of the other architecture, demonstrating the potential of eliminating NMS in the action-spotting task.

\section{More examples of incorrect spotting}

\begin{figure*}[t]
    \centering
    \begin{subfigure}{1.0\linewidth}
        \includegraphics[keepaspectratio,width=\linewidth]{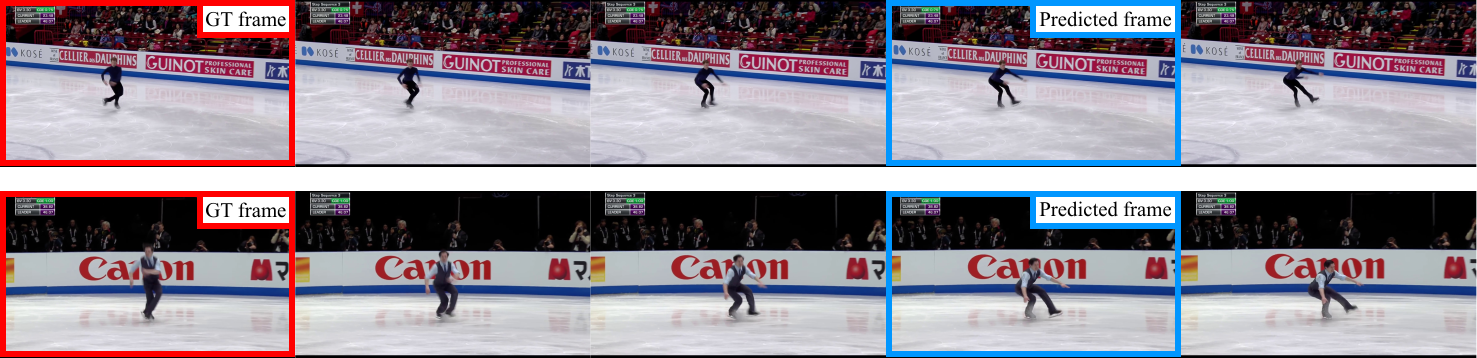}
        \caption{Jump landing}
        \label{fig:offset_weight_addition_jumpland}
    \end{subfigure} \\
    \begin{subfigure}{1.0\linewidth}
        \includegraphics[keepaspectratio,width=\linewidth]{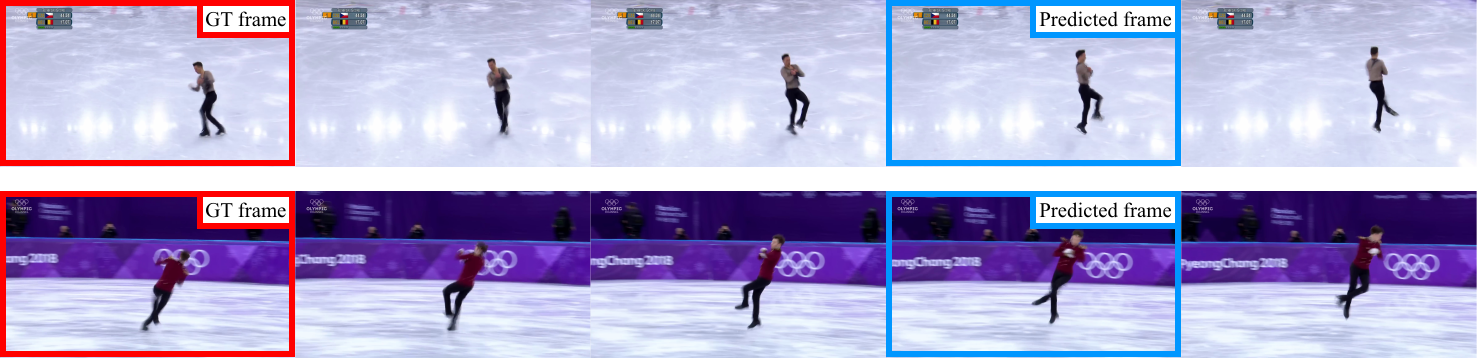}
        \caption{Jump take off}
        \label{fig:offset_weight_addition_jumpoff}
    \end{subfigure} \\
    \begin{subfigure}{1.0\linewidth}
        \includegraphics[keepaspectratio,width=\linewidth]{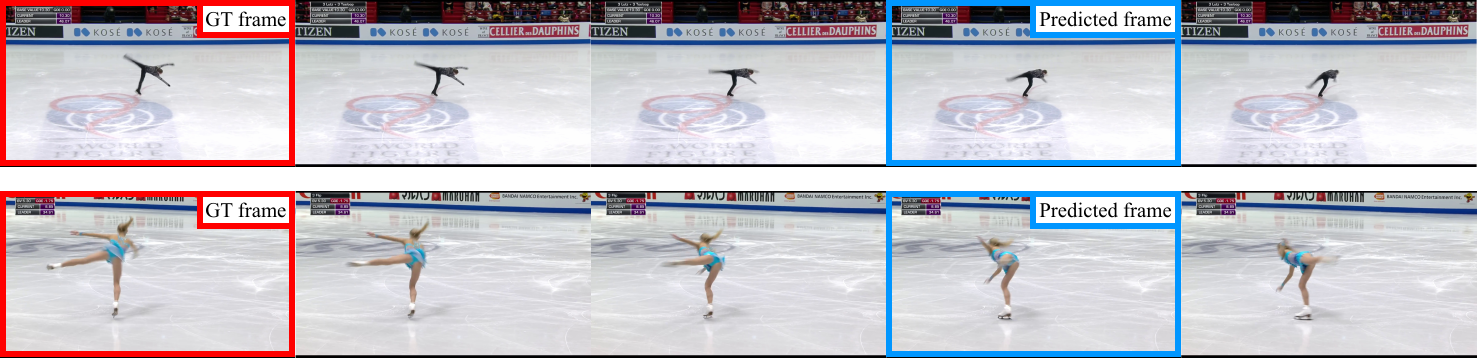}
        \caption{Spin landing}
        \label{fig:offset_weight_addition_spinland}
    \end{subfigure} \\
    \begin{subfigure}{1.0\linewidth}
        \includegraphics[keepaspectratio,width=\linewidth]{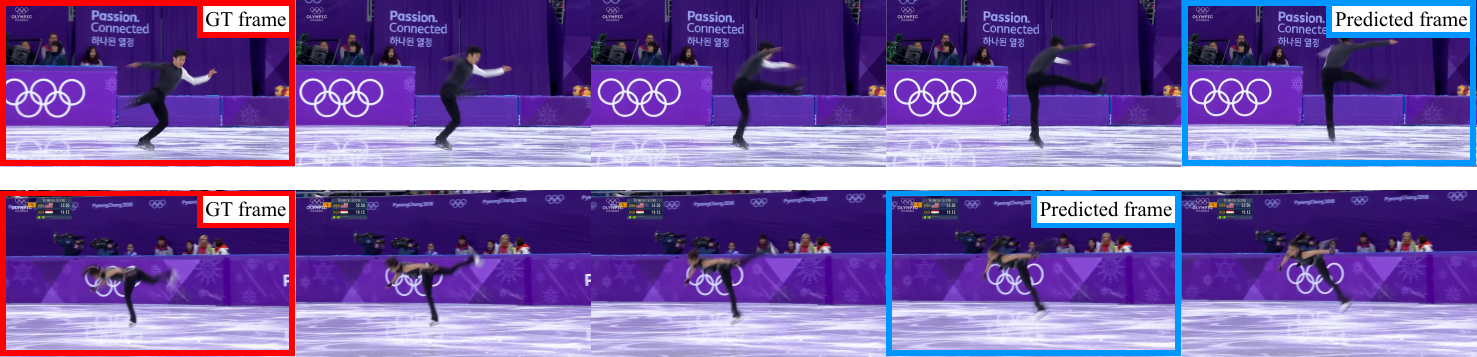}
        \caption{Spin take off}
        \label{fig:offset_weight_addition_spinoff}
    \end{subfigure}
    \caption{Additional example cases where our model recognizes the action classes correctly but identifies frames subsequent to tagged frames as event frames in the test set of the FS dataset.}
    \label{fig:offset_weight_addition}
    \vspace{-1.5em}
\end{figure*}

As outlined in Sec.~{4.5.2} of the main paper, our model tends to recognize actions accurately but identifies frames that are apart from tagged frames as event frames within the FS and FG datasets. Figure~\ref{fig:offset_weight_addition} provides additional examples of the cases from the test set of the FS dataset. In Fig.~\ref{fig:offset_weight_addition_jumpland} and~\ref{fig:offset_weight_addition_spinland}, our model identifies frames where the players' legs are on the ground more apparently compared to the tagged frames for the action classes ``Jump landing" and ``Spin landing." On the other hand, Fig.~\ref{fig:offset_weight_addition_jumpoff} and~\ref{fig:offset_weight_addition_spinoff} illustrate that our model recognizes frames where the players' legs are off the ground more clearly than the tagged frames for the action classes ``Jump take off" and ``Spin take off." As explained in the main paper, our model frequently identifies frames that follow the tagged frames as event frames in the FS dataset because the actions in these subsequent frames are typically more observable compared to the tagged frames. This trend is further supported by the examples shown in Fig.~\ref{fig:offset_weight_addition}.

{
    \small
    \bibliographystyle{ieeenat_fullname}
    \bibliography{main}
}

\end{document}